\pdfoutput=1 
\documentclass{llncs} %

\usepackage{vmargin} 
\setpapersize{A4}
\setmargins{3.5cm}
					 {1.5cm}
           {14.7cm}
           {23.42cm}
           {14pt}
           {1cm}
           {0pt}
           {2cm}

\usepackage{t1enc} 
\usepackage[latin1]{inputenc} 
\usepackage{times}		

\usepackage[pdftex]{graphicx} 
\usepackage{subfigure}
\usepackage{color} 
\usepackage{eso-pic} 
\usepackage{everyshi} 

\usepackage[pdfborder=000,pdftex=true]{hyperref}

\usepackage{booktabs}	  	

\usepackage{ccaption} 
\captionnamefont{\bf\footnotesize\sffamily} 
\captiontitlefont{\footnotesize\sffamily} 
\usepackage{scrpage2} 

\defpagestyle{cb}{%
(\textwidth,0pt)
{\pagemark\hfill\headmark\hfill}
{\hfill\headmark\hfill\pagemark}
{\hfill\headmark\hfill\pagemark}
(\textwidth,1pt)}
{(\textwidth,1pt)
{{\it The University of Nottingham}\hfill Anh Cat Le Ngo}
{Anh Cat Le Ngo\hfill{\it The University of Nottingham}}
{Anh Cat Le Ngo\hfill{\it The University of Nottingham}} 
(\textwidth,0pt)
}

\renewpagestyle{plain}%
	{(\textwidth,0pt)%
		{\hfill}{\hfill}{\hfill}%
	(\textwidth,0pt)}%
	{(\textwidth,0pt)%
		{\hfill}{\hfill}{\hfill}%
	(\textwidth,0pt)}

\begin{document}



\title{Multiscale Discriminant Saliency for Visual Attention}
\titlerunning{Multiscale Discriminant Saliency with Hidden Markov Tree Modeling}
\author{Anh Cat Le Ngo \inst{1} \and Kenneth Li-Minn Ang \inst{2} \\ Guoping Qiu \inst{3} \and Jasmine Seng Kah-Phooi \inst{4}}
\institute{
School of Engineering, The University of Nottingham, Malaysia Campus
\and 
Centre for Communications Engineering Research, Edith Cowan University
\and 
School of Computer Science, The University of Nottingham, UK Campus
\and
Department of Computer Science \& Networked System, Sunway University 
}
\maketitle

\begin{abstract}
The bottom-up saliency, an early stage of humans' visual attention, can be considered as a binary classification problem between center and surround classes. Discriminant power of features for the classification is measured as mutual information between features and two classes distribution. The estimated discrepancy of two feature classes very much depends on considered scale levels; then, multi-scale structure and discriminant power are integrated by employing discrete wavelet features and Hidden markov tree (HMT). With wavelet coefficients and Hidden Markov Tree parameters, quad-tree like label structures are constructed and utilized in maximum a posterior probability (MAP) of hidden class variables at corresponding dyadic sub-squares. Then, saliency value for each dyadic square at each scale level is computed with discriminant power principle and the MAP. Finally, across multiple scales is integrated the final saliency map by an information maximization rule. Both standard quantitative tools such as NSS, LCC, AUC and qualitative assessments are used for evaluating the proposed multiscale discriminant saliency method (MDIS) against the well-know information-based saliency method AIM on its Bruce Database wity eye-tracking data. Simulation results are presented and analyzed to verify the validity of MDIS as well as point out its disadvantages for further research direction.
\end{abstract}

\section{Visual Attention - Computational Approach }
Visual attention is a psychological phenomenon in which human visual systems are optimized for capturing scenic information. Robustness and efficiency of biological devices, the eyes and their control systems, visual paths in the brain have amazed scientists and engineers for centuries. From Neisser \cite{neisser1967} to Marr \cite{marr1976}, researchers have put intesive effort in discovering attention principles and engineering artificial systems with equivalent capability. For last two decades, this research field is dominated by visual saliency principles which proposes an existence of a saliency map for attention guidance. The idea is further promoted in Feature Integration Theory (FIT) \cite{treisman1980} which elaborates computational principles of saliency map generation with center-surround operators and basic image features such as intensity, orientation and colors. Then, Itti et al. \cite{itti1998} implemented and released the first complete computer algorithms of FIT theory \footnote{http://ilab.usc.edu/toolkit/}.
Feature Integration Theory are widely accepted as principles behind visual attention partly due to its utilization of basic image features. Moreover, this hypothesis is supported by several evidences from psychological experiments. However, it only defines theoretical aspects of visual attention with saliency maps, but does not investigate how such principles would be implemented algorithmically. It leaves research field open for many later saliency algorithms \cite{itti1998},\cite{gao2004},\cite{sun2003},\cite{harel2007}, etc. Saliency might be computed as a linear contrast between features of central and surrounding environments across multiple scales in the center-surround operation. Saliency is also modeled as phase difference in Fourier Transform Domain \cite{hou2007}, or saliency at each location depends on statistical modeling of the local feature distribution \cite{sun2003}. Though many approaches are mentioned in long and rich literature of visual saliency, only a few are built on a solid theory or linked to other well-established computational theory. Among the approaches, Neil Bruce's work \cite{bruce2006} nicely established a bridge between visual saliency and information theory. It put a first step for bridging two alien fields; moreover, visual attention for first time could be modeled as information system. Then, information-based visual saliency has continuously been investigated and developed in several works \cite{lengo2012},\cite{lengo2010}, \cite{qiu2007}, \cite{gao2007-a}. The distinguish points between these works are computational approaches for information retrieval from features. The process attracts much interest due to difficulty in estimating information of high dimension data like 2-D image patches. It usually runs into computational problems which can not be efficiently solved due to the curse-of-dimensionality; moreover, central and surrounding contexts are usually defined in ad-hoc manners without much theoretical supports. To encounter the difficulties, Danash Gao et al. has simplified the information extraction step as a binary classification problem with decision theory. Two classes are identified as center and surround contexts then discriminant power or mutual information between features and classes are estimated as saliency values for each location. This formulation of visual saliency approach is named as Discriminant Saliency (DIS) of which underlying principles are carefully elaborated by Gao et al. \cite{gao2009}. Its significant point is estimating information from class distributions given input features rather than from the input features themselves. Therefore, computational load is significantly reduced since only simple class distribution need estimating rather than complex feature distribution.

Spatial features have dominated influence on saliency values; however, scale-space features do have a decisive role in visual saliency computation since center or surround environments are simply processing windows with different sizes. In signal processing, scale-space and spectral space are two sides of a coin; therefore, there is a strong relation between scale-frequency-saliency in visual attention problem. Several researchers \cite{baddeley2006,reinagel1999,parkhurst2002,tatler2005} outlined that fixated regions have high spatial contrast or showed that high-frequency edges allow stronger discrimination between fixateed over non-fixated points. In brief, they all come up with one conclusion increased predictability at high frequencies. Though these studies emphasizes a greater visual attraction to high frequencies (edges, ridges, other structures of images), there are other works focusing on medium frequency. Bruce et al. \cite{bruce2007-a} found that fixation points tend to prefere horizontal and vertical frequency content rather than random position, and these oriented content have more noticeable difference in medium frequencies. More interestingly, choices of frequency range for visual processing may depend on encountering visual context \cite{acck2009}. For example, luminance contrast explained fixation locations better in natural image category and slightly worse in urban scenes category provided that all images are applied low-pass filters as preprocessing steps. Perhaps, that attention system might include different range of frequencies in generating optimal eye-movements. Diversity in spectral space usage means utilization of several different scales in scale-space theory. It can be assumed that both high frequency (small scale) and medium frequency (medium scale) constitutes an ecological relevance and compromise between information requirement and available attentional capacity in the early stage of visual attention when observers are not driven by performing any specific tasks. 

Though multi-scale nature have been emphasized as implicit element of human visual attention, it is often ignored in several visual saliency algorithm. For example, DIS approach \cite{gao2009} considers only one fixed-size window and it may lead to inconsideration of significant attentive features in a scene. Therefor, DIS approach needs constituting under the multi-scale framework to form multiscale discriminant saliency (MDIS) approach. This is the main motivation as well as contribution of this paper which are organized as follows. Section \ref{sec:dis} reviews principles behind DIS \cite{gao2007-a} and focuses on its important assumption and limitation. After that, MDIS approach is carefully elaborated in section \ref{sec:mdis} with several relating contents such as multiple dyadic windows for binary classification problem in subsection \ref{subsec:mcw}, multiscale statistical model of wavelet coefficients in subsection \ref{subsec:msm}, maximum likehood (MLL) and maximum a posterior probability (MAP) computation of dyadic subsquares in subsections \ref{subsec:mlc}, \ref{subsec:map}. Then, all steps of MDIS are combined for final saliency map generation in subsection \ref{subsec:mdis}. Quantitative and qualitative analysis of the proposed method with different simulation modes are discussed in section \ref{sec:exp}; moreover, simulation data of MDIS in comparisons with the well-known information-based saliency method AIM \cite{bruce2006} are presented with a number of interesting conclusions. Finally, main contributions of this paper as well as further research direction are stated in the conclusion section \ref{sec:cls}.
\section{Visual Attention - Discriminant Saliency}
\label{sec:dis}
Saliency mechanism plays a key role in perceptual organization; therefore, recently several researchers attempt to generalize principles for visual saliency. In the decision theoretic point of view, saliency is regarded as power for distinguishing salient and non-salient classes; moreover, discriminant saliency combines classical center-surround hypothesis with derived optimal saliency architecture. In other word, saliency of each image location is identified by the discriminant power of a feature set with respect to the binary classification problem between center and surround classes. Based on decision theory, this discriminant saliency detector can work with variety of stimulus modalities, including intensity, color, orientation and motion. Moreover, various psychophysic property for both static and motion stimuli are shown to be accurately satisfied quantitatively by DIS saliency maps. 

Perceptual systems evolve for producing optimal decisions about the state of surrounding environments in a decision-theoretic sense with minimum probability of error. Beside accurate decisions, the perceptual mechanisms should be as efficient as possible. Mathematically, the problem needs defining as (1) a binary classification of interest stimuli (salient features) against the null hypothesis (non-salient features) and (2) measurement of discriminant power from extracted visual features as saliency at each location in the visual field.  The discriminant power is estimated in classification process with respect to two classes of stimuli: stimuli of interest and null stimuli of all uninterested features. Each location of visual field can be classified whether it includes stimuli of interest optimally with lowest expected probability of error. From pure computational standpoint, the binary classification for discriminant features are widely studied and well-defined as tractable problem in the literature. Moreover, the discriminant saliency concept and the decision theory appear in both top-down and bottom-up problems with different specifications of stimuli of interest \cite{gao2004},\cite{gao2007-a}.

The early stages of biological vision are dominated by the ubiquity of ``center-surround'' operator; therefore, bottom-up saliency is commonly defined as how certain the stimuli at each location of central visual field can be determined against other stimuli in its surround. In other words, ``center-surround'' hypothesis is a natural binary classification problem which can be solved by well-established decision theory. In this problem, classes can be defined as follows.
\begin{itemize}
	\item Center class: observations within a central neighborhood $W_{l}^{1}$ of visual fields location $l$.
	\item Surround class: observations within a surrounding window $W_{l}^{0}$ of the above central region.
\end{itemize}
At each location, likelihood of either hypothesis depends on the visual stimulus, of a predefined set of features X. The saliency at location $l$ should be measured as discriminating power of features $X$ in $W_{l}^1$ against features $X$ in $W_{l}^0$. In other words, discriminant saliency value is proportional to distance between feature distributions of center and surrounding classes. 

Feature responses within the windows are drawn from the predefined feature sets $X$ in a process. Since there are many possible combinations and orders of how such responses are assembled, the observations of features can be considered as a random process, $X(l)=(X_{1}(l),\ldots,X_{d}(l))$ of dimension $d$.  This random process is drawn conditionally on the states of hidden variable $Y(l)$, which is either center or surround state. Feature vector $x(j)$ such that $j\in W_{l}^{c},c\in\{0,1\}$ are drawn from classes $c$ according to the conditional probability density $P_{X(l)}\vert_{Y(l)}(x|c)$ where $Y(l)=0$ for surround or $Y(l)=1$ for center. The saliency of location l, $S(l)$ is equal to the discriminant power of X for the classification of the observed feature vectors. That discriminant concept is quantified by the mutual information between feature, X and class label, Y. 

\begin{eqnarray*}
S(l) & = & I_{l}(X;Y)\\
 & = & \sum_{c}\int p_{X,Y}(x,c)log\frac{p_{X,Y}(x,c)}{p_{X}(x)p_{y}(c)}dx
\end{eqnarray*}

Though binary classification and decision theory makes discriminant saliency computationally feasible, it is only true for low-dimensional data. Computer vision and visual attention need to deal with high-dimensional input images especially when it involves statistics and information theory. As mentioned previously, observations of feature responses $X(l)$ are considered as a random process in $d$-dimensional space. Mutual information estimation in such high-dimensional space encounters serious obstacles due to the curse of dimensionality. As these problems persist, saliency algorithms would never be biologically plausible and computationally feasible. Therefore, discriminant saliency algorithms have to be approximated by taking into account statistical characteristics of natural images as well as mathematical simplification. Dashan Gao and Nuno Vasconcelos have proposed a feasible algorithm for mutual estimation. Mathematically, it can be formulated as follows.
\begin{eqnarray}
I_{l}(X;Y) & = & H(Y)-H(Y\vert X)\\
			 & = & E_{X}(H(Y)+E_{Y\vert X}[logP_{Y\vert X}(c\vert x)]]\\
			& = & E_{X}\left[H(Y)+\sum_{c=0}^{1}P_{Y\vert X}(c\vert x)logP_{Y\vert X}(c\vert x)\right]\\
			& = & \frac{1}{\vert W_{l}|}\sum_{j\in W_{l}}\left[H(Y)+\sum_{c=0}^{1}P_{Y\vert X}(c\vert x(j))log\, P_{Y\vert X}(c\vert x(j))\right]
			\label{eq:dis}
\end{eqnarray}
where $H(Y)=-\sum_{c=0}^{1}P_{Y}(c)logP_{Y}(c)$ is the entropy of classes $Y$ and $-E_{Y\vert X}\left[logP_{Y\vert X}(c\vert x)\right]$ is the conditional entropy of $Y$ given $X$. Given a location l, there are corresponding center $W_{l}^{1}$ and surround $W_{l}^{0}$ windows along with a set of associated feature responses $x(j), j\in W_{l}=W_{l}^{0}\cup W_{l}^{1}$. The mutual information can be estimated by replacing expectations with means of all samples inside the join windows $W_l$. The conditional entropy $H(Y|X)$ can be computed by analytically deriving MAP $P(Y|X)$ given that transformed features are modelled by Generalized Gaussian Distribution (GGD) and only binary classification is considered. Lets name Gao's proposal for discriminant saliency computation as DIS; more details about DIS can be found in their publications \cite{gao2009}\cite{gao2007-a}\cite{gao2007}\cite{gao2008}.

While DIS successfully defines discriminant saliency in information-theoretic senses, its implementation has certain limits. Feature responses are randomly sampled in a single fixed-size window; therefore, it is obviously biased toward objects with distinctive features fitted in that window size. As previous visual attention has confirmed involvement of multi-scales factor in visual attention, DIS needs extension from a fixed-scale process to a multi-scale process. In theory, DIS can be carried out with different size of windows, and this approach certainly produces image responses and saliency values at multiple scales. However, such an approach are not recommended for both computational and biological efficiency. Moreover, it causes high redundancy in saliency values across multiple scales. In order to solve the multi-scale problems systematically, DIS need integration with multiple scale processing techniques such as wavelet .

\section{Multiscale Discriminant Saliency}
\label{sec:mdis}
Expansion from a fixed window-size to multiscale processing is a common problem of algorithms development for computer vision applications. Therefore, there are several framework with multi-scale processing capability, which can be used to develop a so called Multiscale Discriminant Saliency (MDIS). A chosen framework has to include both binary classification and multi-scale processing. In other words, it needs classifying a single image point into two or more separate classes with prior knowledge from other scales. With respect to these requirements, a multiscale image segmentation framework should be a great starting point for MDIS as DIS can be considered as simplification of image segmentation in a sense that it only needs to classify a data point into two classes ( center-or-surround ). DIS only uses this binary classification as intermediate step to measure discriminant power of center-surrounding features, and classification step of DIS does not emphasize on accuracy of segmentation results. 

Typical algorithms employ a classification window of some size in a vague hope that all included pixels are belong to the same class. Obviously, these algorithms and DIS have similar problems with choices of processing window sizes. Clearly, the size of classification is crucial to balance between the classification reliability and accuracy. A large window usually provides rich statistical information and enhance reliability of the algorithm. However, it simultaneously risks including heterogeneous elements in the window and inevitably reduces segmentation accuracy. Appropriate window sizes are equivalently vital for DIS to avoid local maxima in discriminant power as well. If sizes of classification windows are too large or too small, the algorithm risks losing useful discriminative features or being too susceptible to noise. In brief, sizes of processing windows and a number of involved data points have vital impact on both DIS and a segmentation problem.

\subsection{Multiple Classification Windows}
\label{subsec:mcw}
\begin{figure}[!htbp]%
\centering
\includegraphics[width=0.5\columnwidth]{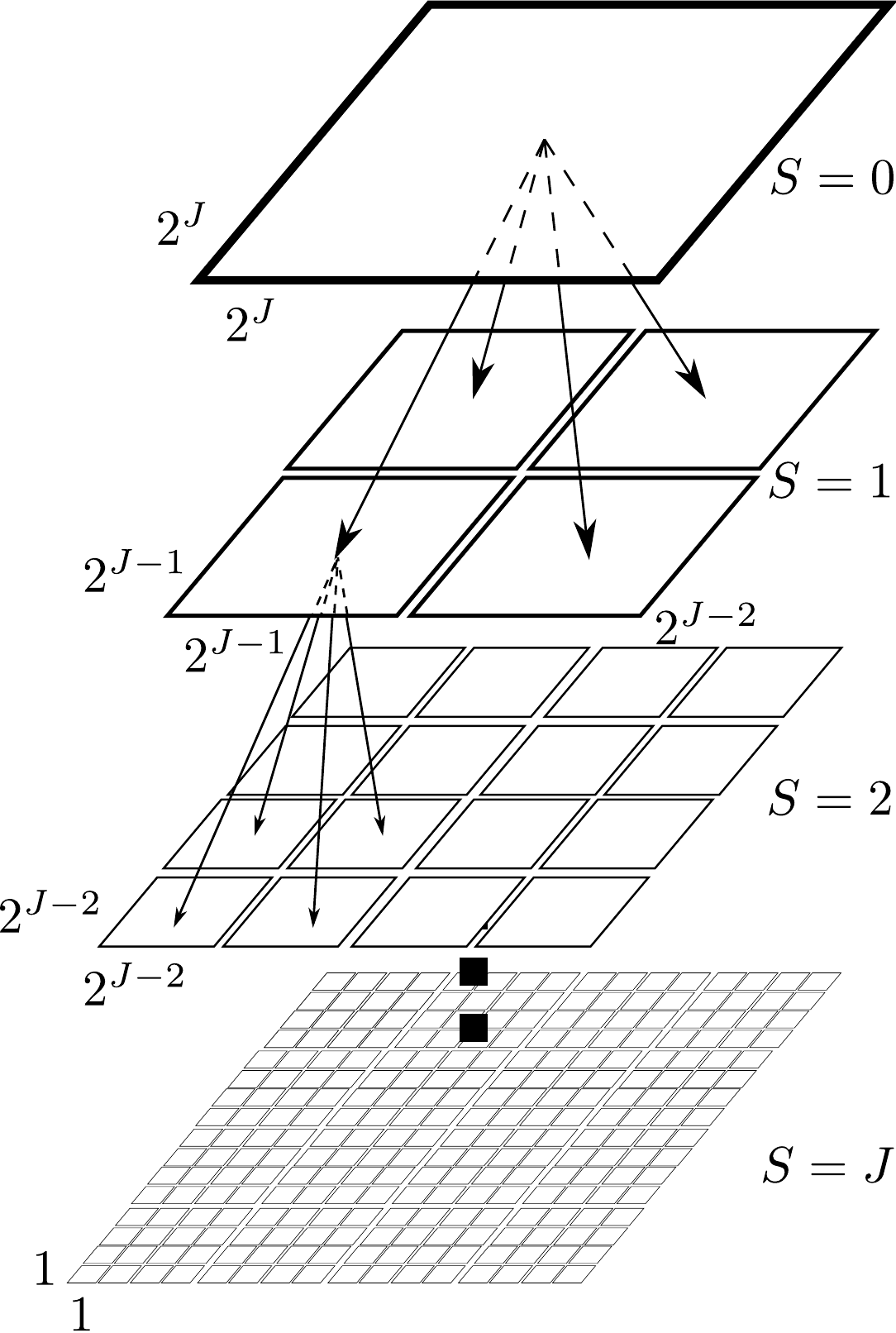}%
\caption{Quad-tree structure}%
\label{fig:quadtree}%
\end{figure}

Multiscale segmentation employs many classification window of different sizes and classifying results are later combined to obtain more accurate segmentation at fine scales. MDIS adapts a similar approach to increase classification efficiency between features of center and surrounding classes. In this paper, dyadic squares ( or blocks ) are implemented as classification windows with different sizes. Lets assume an initial square image $s$ with $2^J$x$2^J$ of $n:=2^{2J}$ pixels, the dyadic square structures can be generated by recursively dividing $x$ into four square sub-images equally. As a result, it has the popular quad-tree structure, commonly employed in computer vision and image processing problems. In this tree structure, each node is related to a direct above parent node while it plays a role of parent node itself for four direct below nodes \ref{fig:quadtree}. Each node of the quad-tree corresponds to a dyadic square, and let denote a tree-node in scale j by $d_{i}^{j}$ whereof i is a spatial index of the dyadic square node. Given a random field image $X$, the dyadic squares are also random fields which are formulated as $D_{i}^{j}$ mathematically. In following sections, we sometime use $D_{i}$ (dropping scale factor $j$) as general randomly-generated dyadic square regardless of scales.

With this predefined structures of classification windows, we will classify each nodes $d_i$ or dyadic squares into either of binary classes center-surround. Using the likelihood of such classification, we can estimate mutual information between features and corresponding labels. Such a mutual information has been proved to quantify discriminant power of central against surrounding features at each location or corresponding discriminant saliency value. This estimation requires multiple pixel pdf models for each class in multiple scales, and these distributions can be learned through wavelet-based statistical models.

\subsection{Multiscale Statistical Model}
\label{subsec:msm}

The complete joint pixel pdf is typically overcomplicated and difficult to model due to their high dimension nature. Unavailability of simple distribution model in practice motivates statistical modeling on transform domain which is often less complex and easier to be estimated. Obviously, joint pixel pdf could be well approximated as marginal pdf of transformed coefficients. Since wavelet transform well-characterizes semantic singularity of natural images (general image category considered in this study), it provides a suitable domain for modeling statistical property of singularity-rich images. 
Natural images are full of edges, ridges other highly structural features as well as textures; therefore, wavelet transforms as multi-scale edge detectors well represents that singularity-rich content in at multiple scales and three different directions. Noted that, only normal discrete wavelet transform (DWT) is considered in this study for the sake of simplicity though concepts can be adapted into other wavelet transforms as well. Henceforth, whenever wavelet transform is mentioned, it should refer to DWT instead of stating otherwise. As multi-scale edge detectors, responses of wavelet transform overlying a singularity are large wavelet coefficients while wavelets overlying a smooth region yield small coefficients. Deploying thresholds over wavelet coefficients leads to binary classification of singularity against non-singularity features. Moreover, statistical model of images can be simplified under "`restructure"' singularity representation in multiscale manner.
Quad-tree structure of dyadic squares in pixel domain can be mirrored in wavelet decomposition since four wavelet coefficients at a given scale nest inside one at the next coarser level. For example, Haar wavelet coefficients at each quad-tree nodes are generated by Harr wavelet transform of the corresponding dyadic image square. The marriage between singularity detector and multi-scale quad-tree structure implies the singularity property at each spatial location persists through scales along the branches of the quad-tree.
The singularity characterization along scales makes the wavelet domain well-suited for modeling natural images. In fact, statistical modeling of wavelet coefficients have quite comprehensive literature; here we concentrate on the hidden Markov tree (HMT) of Crouse, Nowak and Baraniuk \cite{crouse1998}. In consideration of both marginal and joint wavelet coefficient statistics, the HMT model introduces a (hidden) state variable of either "`large"' or "`small"' to each wavelet coefficients. Then, the marginal density of wavelet coefficients is modeled as a two-density Gaussian mixture in which a "`large"' state refers to a large variance Gaussian and a "`small"' state represents small variance Gaussian distribution. This Gaussian mixture closely matches marginal statistics observed in natural images \cite{pesquet1996},\cite{abramovich1998}, \cite{chipman1997}. With the HMT, persistence of large or small coefficients are captured across scales using Markov-1 chain. It models dependencies between hidden states across scale in a tree structure, parallel to that of wavelet coefficients and dyadic squares. Combining both the Gaussian Mixture Model (GMM) and Markov State Transition into a vector $\cal{M}$, the wavelet HMT is able to approximate the overall joint pdf of the wavelet coefficients $\mathbf{W}$ by a high-dimensional but highly structured Gaussian mixture models $f(\mathbf{w} \vert \cal{M})$. 
Highly structural nature of wavelet coefficients allows efficient implementation of HMT-based processing. The HMT model parameters $\cal{M}$ can be learned through the iterative expectation and maximization (EM) algorithm with cost ${\cal O}(n)$ per iteration \cite{crouse1998} or predefined for a particular image category \cite{romberg2001}. After the parameters $\cal{M}$ are estimated by EM algorithm, we need to compute statistical characteristics of wavelet coefficients given the wavelet transform $\mathbf{\tilde{w}}$ of a test image $\mathbf{\tilde{x}}$ and a set of HMT parameters ${\cal M}$. It is a realization of the HMT model in which computation of the likelihood $f(\mathbf{\tilde{\omega}\vert{\cal{M}})}$that $\mathbf{\tilde{\omega}}$ requires only a simple ${\cal O}(n)$ unsweep through the HMT tree from leaves to root \cite{crouse1998}.
Conveniently, HMT models wavelet coefficients in a tree parallel to quad-tree structures of wavelet coefficients and dyadic squares. Therefore, statistical behavior of each dyadic square $\mathbf{\mathbf{\tilde{d}}_{i}}$ can also be approximately modeled by a HMT branch rooted at node $i$. As mentioned earlier, maximum likelihood of the sub-tree ${\cal T}_i$ is computed simply by up-sweeping from corresponding leave-nodes at scale $S = J$ to the root node at scale $S = j$. If the "upsweeping" operation is done from tips to toes, from leave nodes to root node at scale "S = 0", we could find out likelihood probability of the whole image. The partial likelihood calculations at intermediate scales $j$ of the HMT tree is denoted as $f(\mathbf{\tilde{d}}_{i}\vert{\cal M})$, and it is also statistical behavior of the corresponding dyadic sub-square under the HMT model as well. 

The above model leads to a simple multi-scale image classification algorithm. Supposed that there are two main classes: center and surround environment $c\in \{1,0\}$, we have specified or trained an HMT tree for each class with parameters ${\cal M}_{c}$. When the above likelihood calculation is deployed on each node of the HMT quad-tree given the wavelet transform $\tilde{\mathbf{\omega}}$ of an image $\mathbf{\tilde{x}}$. For each node of the tree, HMT yields the likelihood $f(\tilde{\mathbf{d_{i}}|{\cal M}_{c}),}c\in\{1,0\}$ for each dyadic sub-image $d_{i}$. With the multiscale likelihoods at hand, we can choose the most suitable class $c$ for a dyadic sub-square $\mathbf{\tilde{d_i}}$ as follows.
\[
\hat{c}_{i}^{ML}:=argmax_{c \in \{1,2\}}f(\mathbf{\tilde{d}_{i}}|{\cal M}_{c})
\] 
 The most likely label $\hat{c}_{i}^{ML}$ for each dyadic sub-square $\mathbf{\tilde{d_{i}}}$ can be found by simple comparison of likelihood among available classes. Moreover, likelihood of the whole tree just needs ${\cal O}(n)$ operations for an n-pixel image. 

\subsection{Multiscale Likelihood Computation}
\label{subsec:mlc}
For a given set of HMT model parameters ${\cal M}$, it is straight forward to compute the likelihood $f(\mathbf{\tilde{w}}\vert {\cal M})$ by upsweeping from leaves nodes to current node in a single sub-band branch \cite{crouse1998}. Moreover, likelihoods of all dyadic squares of the image can be obtained simultaneously in upsweep operations along the tree as well due to compatibility between HMT , wavelet decompositions and dyadic square quad-trees. 

To obtain the likelihood of a sub-tree ${\cal T}_i$ of wavelet coefficients rooted at $w_i$, we have deployed wavelet HMT trees and learn parameters $\Theta$ for multi-scale analysis \cite{crouse1998}. The conditional likelihood $\beta_i(m):=f({\cal T}_i|S_i = m,\Theta)$ can be retrieved by sweeping up to node $i$ (see \cite{crouse1998}); then, the likelihood of the coefficients in ${\cal T}_i$ can be computed as.
\begin{equation}
f({\cal T} \vert \Theta) = \sum_{m=S,L} \beta_i(c)p(S_i = c \vert \Theta)
\label{eq:mlc1}
\end{equation}
with $p(S_i=c\vert\Theta)$ state probabilities can be predefined or obtained during traning \cite{romberg2001}. 

From the previous discussion about relations between wavelet coefficients and dyadic squares, it is obvious that each dyadic square $\mathbf{d_i}$ is well represented by three sub-bands $\{ {\cal T}_i^{LH}, {\cal T}_i^{HL}, {\cal T}^{HH}\}$. All above likelihood can be estimated by upsweeping operations in their corresponding trees independently. In DIS approach , Gao \cite{gao2009} have also assumed that correlation between feature channels would not affect discriminant powers; moreover, there is de-correlation characteristics of DWT. Therefore, the likelihood of a dyadic square is formulated as product of three independent likelihoods of three wavelet sub-bands at each scale.
\begin{equation}
	f(d_i \vert {\cal M} ) = f({\cal T}_{i}^{LH}\vert \Theta^{LH})f({\cal T}_{i}^{HL}\vert \Theta^{HL})f({\cal T}_{i}^{HH}\vert \Theta^{HH})
\label{eq:mlc2}
\end{equation}
Using the above equation, the likelihood can be computed for each dyadic square down to 2x2 block scale. Noteworthy, sub-band $HH$ of the wavelet transform is not utilized in our computation since it is low-passed approximation of original images. Therefore, it is vulnerable to pixel brightness and lightning conditions of scenes. Since dealing with shades, lightning changes is out of this paper's scope, the final $HH$ sub-band is discarded. The above simple formulation of likelihood is usually employed in block-by-block classification or "`raw"' classification since it does not exploit any possible relationship at different scales. Therefore, classification decisions between classes (center and surround) are lack of inheritance across dyadic scales since likelihood values are estimated independently at each scale. There still remains a problem of integrating classification processes across all scales or at least the direct coarser scale.
\subsection{Multiscale Maximum a Posterior}
\label{subsec:map}
In the previous section, binary classification between between two states have been realized under the wavelet Hidden Markov Tree model \cite{crouse1998}. It bases on comparison of likelihood given a system parameters. However, realization of DIS and MDIS, the equation \ref{eq:dis}, needs a posterior probability $p(c_{i}^{j}|d^j)$ given $c_{i}^{j}$ and $d^j = d_{i}^{j}$ are class labels and features of an image at a dyadic scale $j$ and location $i$. 

In order to estimate the MAP $p(c_{i}^{j}|d^j)$, we need to employ Bayesian approach and capture dependencies between dyadic squares at different scales. Though many approximation techniques \cite{cheng1997},\cite{cheng1998},\cite{Li2000},\cite{bouman1994} are derived for a practical MAP, the Hidden Markov Tree (HMT) by Choi \cite{Choi2001} is proven to be a feasible solution. Choi introduces hidden label tree modeling instead of estimating joint probability of high-dimensional dyadic squares. Due to strong correlation between the considered square and its parents as well as their neighbors, the class labels of these adjacent squares should affect its class label decision. For example, if parent squares belongs to the center class, their sub-square most likely belongs to the same class as well; neighbor squares holds similar influences. 

A possible solution for modeling these relations between squares is using a general probabilistic graph \cite{crouse1998}; however, the complexity exponentially increases with number of neighborhood nodes. Choi \cite{Choi1998} proposes alternative simpler solution based on context-based Bayesian approach. For the sake of simplicity, causal contexts are only defined by states of the direct parent node and its 8 intermediate neighbors. Lets denote the context for $D_i$ as $\mathbf{v_i} \equiv [v_{i,0},v_{i,1},\ldots,v_{i,8}]$ where $v_{i,0}$ refers to context from a direct parent node, the other contexts from neighboring sub-squares. The triple $\mathbf{v_i} \rightarrow C_i \rightarrow \mathbf{D_i}$ forms a Markov-1 chain, relating prior context $\mathbf{v_i}$ and node features $\mathbf{D_i}$, to classify labels $C_i$. Moreover, class labels of parent nodes and its neighbors $\mathbf{v_i}$ are chosen as discrete values, then it simplifies the modeling considerably. Given prior context, independence can be assumed for label classification at each node; therefore, it is allowed to write.
\begin{equation}
p(\mathbf{c^j} \vert \mathbf{v^j}) = \prod_{i}{p(c_{i}^{j}) \vert \mathbf{v_{i}^{j}}}
\label{eq:map1}
\end{equation}
The property of Markov-1 chain assumes that $\mathbf{D_i}$ is independent from $\mathbf{v_i}$ given $C_i$; therefore, the posterior probability of classifying $\mathbf{c^j}$ given $\mathbf{d^j,v^j}$ is written as follows.
\begin{equation}
p(\mathbf{c^j \vert d^j,v^j}) = \frac{f(\mathbf{d^j \vert c^j})p(\mathbf{c^j \vert v^j})}{f(\mathbf{d^j \vert v^j})}
\label{eq:map2}
\end{equation}
As independence is assumed for label decision from classifying processes, it yields. 
\begin{equation}
p(\mathbf{c^j \vert d^j,v^j}) = \frac{1}{f(\mathbf{d^j \vert v^j)}} \prod_{i}{f(\mathbf{d^j} \vert c^j) p(c^j \vert \mathbf{v^j})}
\label{eq:map3}
\end{equation}
and the marginalized context-based posterior
\begin{equation}
f(c_{i}^{j} \vert \mathbf{d^j,v^j}) \propto f(\mathbf{d_{i}^{j}} \ c_{i}^{j})p(c_{i}^{j} \vert \mathbf{v_{i}^{j}})
\label{eq:map4}
\end{equation}
It greatly simplifies MAP posterior estimation since it no longer needs to deal with joint prior conditions of features and contexts. Rather than that, it only need to obtain two separated likelihood of the dyadic square given the class value $C_i$, $f(\mathbf{d_{i}}^{j} \vert c_{i}^{j})$ and prior context provided through $\mathbf{v_i}$, $p(c_{i}^{j} \vert \mathbf{v_{i}^{j}})$. 

While it is straightforward to retrieve the likelihood $f(\mathbf{d_{i}}^{j} \vert c_{i}^{j})$ by up-sweeping operations with given HMT model parameters at each scale, the complexity of prior context estimation greatly depends on the context choice. Though general context may give better prior information for classification, it also greatly complicates modeling and it is difficult to summarize information conveyed by $\mathbf{v_{i}^{j}}$ as well. In other words, we run on the risk of context dilution, especially with insufficient training data \cite{cheng1997},\cite{cheng1998},\cite{crouse1998}.

To balance simplicity and generalization of prior information, we will employ a simple context structure inspired by the hybrid tree model \cite{bouman1994} for context-labeling tree. Instead of including all neighboring sub-squares, the simplified context only involves labels from the parent square $C_{\rho(i)}$ and majority vote of the class labels from neighboring squares $C_{{\cal N}_{i}}$. Since there are only two class labels $N_c := {0,1}$, the prior context $\mathbf{v_i} := \{ C_{\rho(i)} , C_{{\cal N}_{i}} \}$ can only been drawn from $N_{c}^2 = 4$ different values $\{0,0\},\{0,1\},\{1,0\},\{1,1\}$. While the choice of such simple context is rather ad-hoc, it provides sufficient statistic for demonstrating the effectiveness of multi-scale decision fusion. Another advantage of the context structure simplification is that only few training data are required for probability estimation. 

Any decision about labels at a scale $j$ depends on prior information of labels on a scale $j-1$; therefore, we can maximize MAP, in the equation \ref{eq:map5}, in multi-scale coarse-to-fine manner by fusing the HMT likelihoods $f(\mathbf{d_i} \vert c_i)$ given the label tree prior $p(c_{i}^{j} \vert \mathbf{v_i})$. That fusion will pass down MAP classification decisions through scales to enhance across-scale coherency. Moreover, posterior probability of a class label $c_i$ given features and context are computed and maximized coherently across multiple scale.

\begin{equation}
	\hat{c_i}^{MAP} = argmax_{c_{i}^{j} \in {0,1}} f(c_{i}^{j} \vert \mathbf{d^j,v^j})
\label{eq:map5}
\end{equation}

\subsection{Multiscale Discriminant Saliency}
\label{subsec:mdis}

The core idea of discriminant saliency is discriminant power between two classes center and surrounds. Though the discriminant power is measured by sample means of mutual information, the underlying mechanism is making use of difference between generalized gaussian distributions (GGD) given either center label or surrounding label. Since GGD of wavelet coefficients usually have zero-mean, it is well-characterized with only variance parameters. Dashan Gao \cite{gao2009} tries to estimate the scale parameter or variance of GGD (see section 2.4 \cite{gao2009} for more details) by the maximum a probability process.
\begin{equation}
\hat{\alpha}^{MAP} = \left[ \frac{1}{{\cal K}} \left( \sum_{j=1}^{n} \vert x(j) \vert^\beta + \nu \right) \right]^{\frac{1}{\beta}}
\label{eq:mdis1}
\end{equation}
The above MAP is later used for deciding whether a sample point or a image data point belongs to the center or surround class (see \cite{gao2009} for a detailed proof and explanation). Therefore, the more difference between MAP estimation of the center's variance parameter $\alpha_1$ and the surround's variance parameter $\alpha_0$, the more discriminant power is for classifying interest versus null hypothesis. 

The idea of modeling wavelet distributions with multiple classes' variances can be realized by Gaussian Mixture Model (GSM) \cite{pesquet1996},\cite{chipman1997} as well. In binary classification problem with only two classes, there are mixtures of two states with Gaussian Distribution (GD) of distinguishing variances. It is reasonable to name "`large"' and "`small"' states according to their comparison in terms of variance values. Now the only difference between GSM models and Gao's proposal \cite{gao2007-a} are whether GD or GGD should be used. Though GGD is more sophisticated with distribution shape parameter $\beta$, several factors support validity of simple GD modeling given the class labels as hidden variables. Empirical results from estimation have shown that the mixture model is simple yet effective \cite{pesquet1996},\cite{bouman1994}. Modeling wavelet coefficients with hidden classes of "`large"' and "`small"' variance state are basic data models in Wavelet-based Hidden Markov Model (HMT) \cite{crouse1998}. With wavelet HMT, image data are processed in coarse-to-fine multi-scale manner; therefore, MAP of a state $C_{i}^{j}$ given input features from a sub-square $D_{i}^{j}$ can be inherently estimated across scales $j = {0,1,\ldots,J}$. More details about this multi-scale MAP estimation by wavelet HMT are discussed in the previous sections \ref{subsec:map}. Combination MAP estimation, the equation \ref{eq:map4}, and mutual information estimation, the \ref{eq:dis}, the equation \ref{eq:dis} yields a formulation for multiscale DIS.
\begin{equation}
I_{i}^{j} (C^{j};\mathbf{D^{j}}) = H(C^j)+\sum_{c=0}^{1} P_{C^j \vert \mathbf{D^j}}(c_{i}^{j} \vert \mathbf{d^j}) log P_{C^j \vert \mathbf{D^j}}(c_{i}^{j} \vert \mathbf{d^j})
\label{eq:mdis2}
\end{equation}
where $H(C^j) = - p(C^j) log(p(C^j))$ is entropy estimation of classes across the scale $j$, and the posterior probability can be estimated by modeling wavelet coefficients in HMT frameworks. This matter has been discussed in previous sections; therefore, it is not repeated here. The equation \ref{eq:mdis2} yields discriminant power across multiple scales; meanwhile a strategy is needed for combining them across scales. In this paper, a simple maximum rule is applied for selecting discriminant values from multiple scales into a singular discriminant saliency map at a sub-square $d_i$. 
\begin{equation}
I_{i} (C|\mathbf{D}) = max \left( I_{i}^{j} (C^{j};\mathbf{D^{j}}) \right)
\label{eq:mdis3}
\end{equation}

\section{Experiments \& Discussion}
\label{sec:exp}
In the light of decision theory, saliency maps are considered as binary filters applied at each image locations. According to a certain saliency threshold, each spot can be labeled as interesting or uninteresting. If a binary classification map is considered as saliency map which leads to visual performance of human beings, it would be a significant undervaluation of human visual attention system. Psychology experiments shows much better capacity of biologically plausible visual attention system than that of binary classification maps. Therefore, our proposed method just use binary classification between center-surround environment as an intermediate step to develop information-based saliency map. At each location, mutual information between distributions of classes and features shows strength of discriminant power between interesting vs non-interesting classes given input dyadic sub-squares. From discrete binary values, the saliency representation has continuous ranges of discriminant power. Inevitably, the generated saliency maps become more correlated to the results of human visual attention maps.

Besides appropriate saliency representation, we also need reliable ground-truth to compare with those maps. As our research purpose is deepening knowledge about relationship between multi-scale discriminant saliency approaches and human visual attention, the ground truth data must be gotten from psychological experiments in which human subjects are tested with different natural scenes. Moreover, the research scope only focuses at bottom-up visual saliency, the early stage of visual attention without interference of subjects' prior knowledge and experiences. Human participants should be naive about purposes of experiments and should not know contents of displaying scenes in advance. After these prerequisites are satisfied, human responses on each scene can be accurately collected through eye-tracking equipments. It records collection of eye-fixations for each scene, and these raw data are basic form of ground truths for evaluating efficiency of saliency methods.

Assumed that ground-truth data are available, quantitative methods can be applied for evaluation of MDIS Saliency results. In an effort of standardizing evaluating process of evaluating saliency methods, we only utilize one of the most common and accessible database and evaluation tools in visual attention fields. In regards of available database and ground-truths, Niel-Bruce database \cite{bruce2006} is certainly the most popular dataset used in information-based saliency studies. While proposing his An InfoMax (AIM) saliency approach, the first information-based visual saliency, Bruce simultaneously releases his testing database as well. The reasonably small database with 120 different color images which are tested by 20 different subjects. Each object observes displayed image in random order on a 12 inch CRT monitor positioned 0.75 from their location for 4 seconds with a mask between each pair of images. Importantly, no particular instructions are given except observing the image. Above brief description clarified validity of this database. When setting up our simulations on this database, AIM method, of which codes are released along, is included as referenced method. We compare our proposed saliency method MDIS against AIM in terms of performance, computational load, etc. Though DIS \cite{gao2007-a} is the closest approach to our proposed MDIS, implementation from the author is not available for comparison. Meanwhile, AIM also derives saliency value from information theory with slightly different computation, self-information instead of mutual-information in MDIS or DIS. Therefore, it would be considered the second best as referenced method for our later evaluation of MDIS.

As valid database is set, proper numerical tools are necessary for simulation data analysis. In regards of fairness and accuracy of the evaluation, we employ a set of three measurements LCC, NSS, and AUC recommended by Ali Borji et al. \cite{borji2012-c} since evaluation codes can be retrieved freely from the authors' website \footnote{https://sites.google.com/site/saliencyevaluation/}. Three evaluation scores are used for analyzing a method to ensure the reliability of qualitative conclusion and any conclusion is free from the choice of metric. First linear correlation coefficient (LCC) measures the strenght of linear relationship between two variables $CC(G,S)=cov(G,S)$, where G and S are the standard deviation of ground-truths and corresponding saliency maps. LCC values variates from -1 to +1 while the correlation changes from total inverse to perfect linear relation between G and S maps. While LCC measures matching between saliency maps and eye-fixation maps as a whole, normalized scanpath saliency (NSS) treats eye-fixations as random variables which are classified by proposed saliency maps. The measurement is an average of saliency values at human eye positions according to saliency approaches. NSS values are in between 0 and 1. $NSS = 1$ indicates saliency values at eye-fixation locations are one standard deviation above average, while $NSS = 0$ indicates no better performance of saliency maps than randomly generated maps. The previous two measurements deals with saliency maps directly while the last quantitative tool AUC utilized saliency maps as binary classification filters with various thresholds. By regularly increasing/decreasing threshold values over the range of saliency values, we can have a number of binary classifying filters. Then, deploying these filters on eye-fixation maps produce several true positive rates (TPR) and false positive rate (FPR) as vertical axis and horizontal axis values of Receiver Operating Characteristics (ROC) curve. Area under curve (AUC) is a simple quantitative measurement to compare ROC of different saliency approaches. Perfect AUC prediction means a score of 1 while 0.5 indicates by chance-performance level.

As mentioned previously, AIM is chosen as the referenced information-based saliency method. It is chosen due to the well-established reputation as well as freely accessible code and experiment database \footnote{http://www-sop.inria.fr/members/Neil.Bruce/}. Due to multiscale natures of the proposed MDIS, saliency maps for each dyadic-scale levels can be extracted as well as the final MDIS saliency, integrated across scales by the equation \ref{eq:mdis3}. The availability of saliency maps at multiple scales as well as combined one allows evaluation of discriminant power concept for saliency in scale-by-scale manner or on the whole. We denote integrated MDIS as HMT0; while separated saliency maps are named as HMT1 to HMT5 in accordance with coarse-to-fine order. By examining MDIS in different aspects, we would observe its effectiveness in predictions of eye-fixation points and how selection of classifying window sizes might affect its performance. In addition, comparisons against AIM would contribute a general view how MDIS performs against a well-known information-based saliency method. 

In our proposed saliency techniques, Hidden Markov Tree (HMT) plays a key role of modeling statistical properties of images. It extracts model parameters at each scale by considering distribution of feature given hidden variables (center-surround labels in our method). Therefore, training is a necessary step before model parameters can be approximated in trained Hidden Markov Model (THMT). However, training steps are not necessarily needed if data are restricted in a specific category; for example, natural images in this paper. Romberg et al. \cite{romberg2001} have studied this issue and proposed a Universal Hidden Markov Tree (UHMT) for the natural image class. In other words, model parameters can be fixed without any training efforts; the approach would greatly reduce computational load for MDIS. However, a necessary evaluation step needs carrying out to compare performances of UHMT and THMT in terms of LCC, NSS, AUC and TIME ( the computational time ). Noteworthy that, we will use UHMT or THMT instead of HMT alone when representing experimental data in order to signify which tree-building method is employed.

In quantitative method, general ideas can be drawn about how the proposed algorithms perform in average. However, such evaluation method lacks of specific details about successful and failure cases. The simulation result has been averaged out in quantitative methods. In an effort of looking for pros and cons of the algorithm, we perform qualitative evaluation for saliency maps generated by MDIS in multiple scales. Furthermore, AIM saliency maps are generated and compared with MDIS maps to discover advantages and disadvantages of each method.

\subsection{Quantitative Evaluation}
After general review of how simulations are setup and evaluated in the previous section, following are data representation and analysis of the conducted experiments. In this paper, five dyadic scales are deployed for any HMT training and evaluation; therefore, we have simulation modes from (U)THMT1 to (U)THMT5 of MDIS depending on whether training stages is deployed (THMT) or universal parameters are used (UHMT). Saliency maps could be combined according to the maximization of mutual information rule, the equation \ref{eq:mdis3}, to form two (U)THMT0 modes for saliency maps. AIM is involved in the simulations as benchmarking approach, and LCC, NSS, AUC and TIME are chosen as numerical evaluation tools. Below are two tables of simulation results. Table \ref{tab:uhmt} shows simulation data of all universal HMT modes while table \ref{tab:thmt} summarizes data of all trained HMT modes .

\begin{table}[!htbp]
\caption{UHMT - MDIS - Experiment Data}
\begin{center}
\resizebox{\columnwidth}{!}{
\begin{tabular}{|l|c|c|c|c|c|c|c|}
\hline
Observations & \multicolumn{1}{l|}{UHMT0} & \multicolumn{1}{l|}{UHMT1} & \multicolumn{1}{l|}{UHMT2} & \multicolumn{1}{l|}{UHMT3} & \multicolumn{1}{l|}{UHMT4} & \multicolumn{1}{l|}{UHMT5} & \multicolumn{1}{l|}{AIM} \\ \hline
LCC & 0.01434 & \texttt{-0.00269} & 0.01294 & 0.01349 & \textbf{0.01604} & 0.00548 & 0.01576 \\ \hline
NSS & 0.21811 & 0.19772 & 0.27819 & 0.32868 & \textbf{0.42419} & \texttt{0.13273} & 0.12378 \\ \hline
AUC & \textbf{0.89392} & \texttt{0.53862} & 0.60520 & 0.69065 & 0.83615 & \textbf{0.89234} & 0.72275 \\ \hline
TIME(s) & 0.39617 & 0.39617 & 0.39617 & 0.39617 & 0.39617 & 0.39706 & 50.41714 \\ \hline
\end{tabular}
}
\end{center}
\label{tab:uhmt}
\end{table}

\begin{table}[!htbp]
\caption{THMT - MDIS - Experiment Data}
\begin{center}
\resizebox{\columnwidth}{!}{
\begin{tabular}{|l|c|c|c|c|c|c|c|}
\hline
Observations & \multicolumn{1}{l|}{THMT0} & \multicolumn{1}{l|}{THMT1} & \multicolumn{1}{l|}{THMT2} & \multicolumn{1}{l|}{THMT3} & \multicolumn{1}{l|}{THMT4} & \multicolumn{1}{l|}{THMT5} & \multicolumn{1}{l|}{AIM} \\ \hline
LCC & \textbf{0.02382} & \textbf{0.02582} & 0.01156 & 0.01604 & 0.01143 & \texttt{0.00512} & 0.01576 \\ \hline
NSS & \textbf{0.48019} & 0.38096 & 0.31855 & 0.32491 & \texttt{0.29662} & 0.36932 & 0.12378 \\ \hline
AUC & \textbf{0.88357} & \texttt{0.60922} & 0.64633 & 0.71972 & 0.81192 & \textbf{0.89532} & 0.72353 \\ \hline
TIME(s) & 2.32734 & 2.32734 & 2.32726 & 2.32726 & 2.32726 & 2.32726 & 50.41714 \\ \hline
\end{tabular}
}
\end{center}
\label{tab:thmt}
\end{table}

\begin{figure}[!htbp]
	\centering
	\subfigure[UHMT - MDIS]{\includegraphics[width=0.32\columnwidth]{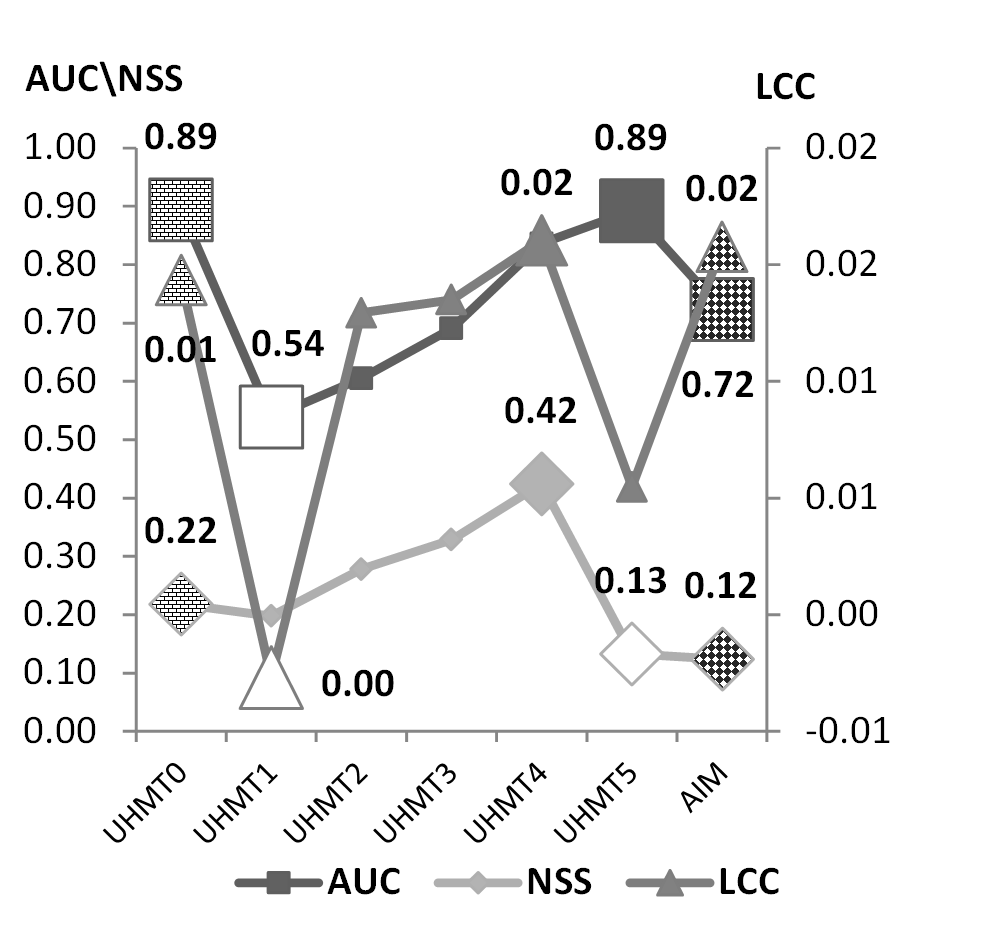}\label{fig:uhmt}}		
	\subfigure[THMT - MDIS]{\includegraphics[width=0.32\columnwidth]{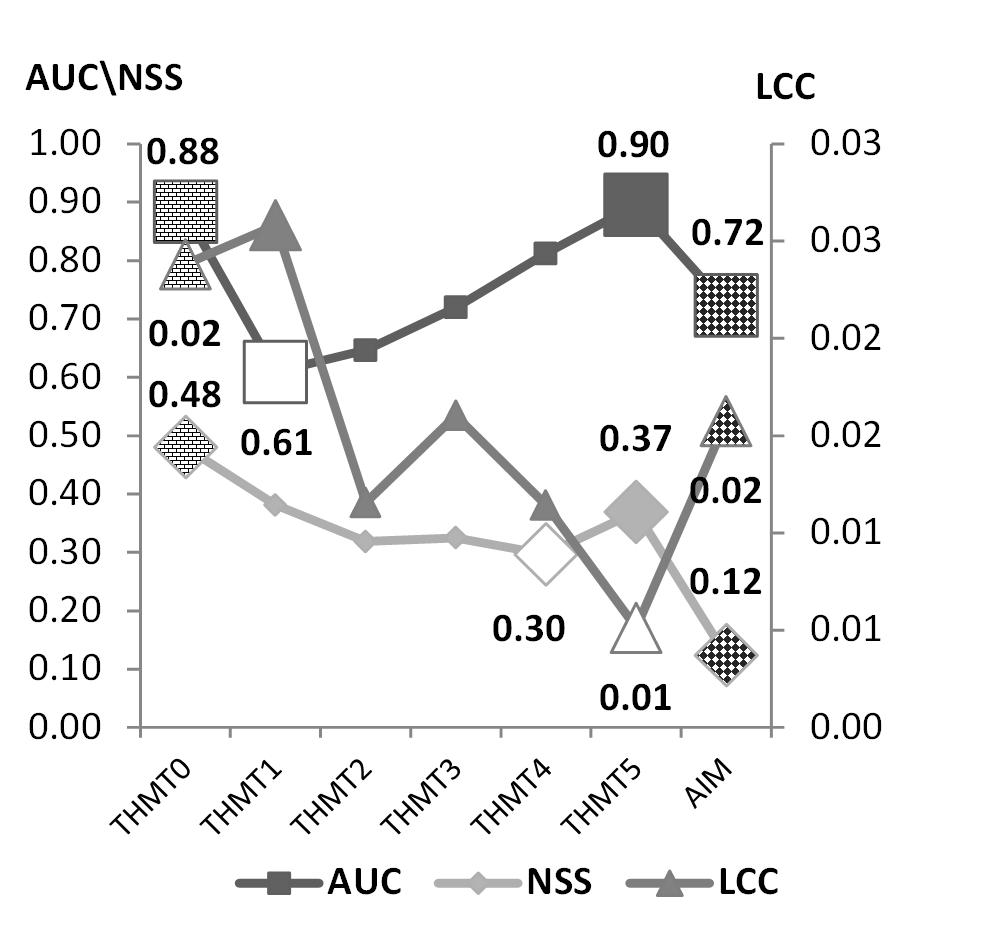}\label{fig:thmt}}		
	\subfigure[TIME]{\includegraphics[width=0.32\columnwidth]{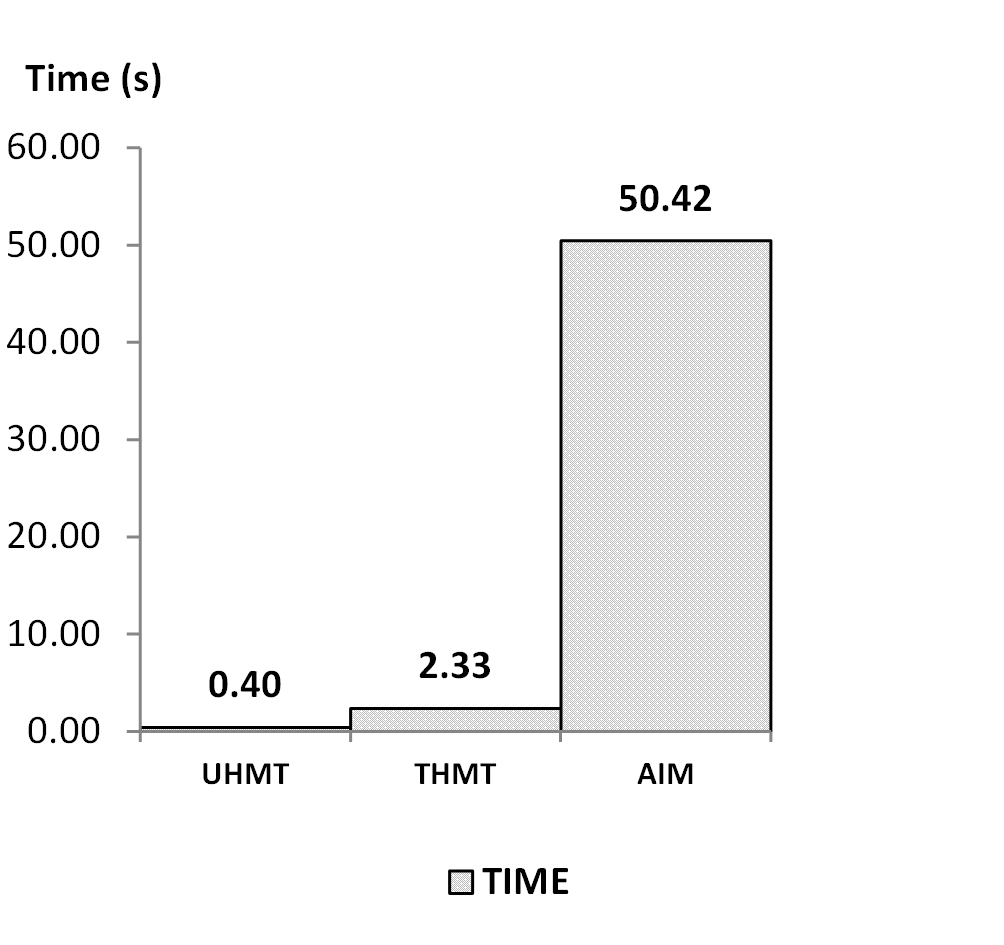}\label{fig:time}}		
	\caption{Performance of UHMT-MDIS, THMT-MDIS in AUC,NSS,LCC and TIME}
	\label{fig:hmtexpres1}
\end{figure}

Looking at TIME rows of both tables \ref{tab:uhmt} and \ref{tab:thmt}, we present necessary processing time for each method or each mode. Generally, computational loads, proportional to processing time, of all modes in either UHMT or THMT rows are almost similar since the parameters of full-depth Hidden Markov Tree need estimating before computation of saliency values for each scale. In comparison between UHMT and THMT in terms of processing time, UHMT is faster than THMT as UHMT uses predefined parameters for HMT and MDIS computation instead of adapting HMT to each image. When comparing both UHMT and THMT modes of MDIS with AIM, our proposed methods are much faster than AIM. The well-known AIM directly estimate self-information from high-dimensional by ICA algorithm while MDIS statistically models two hidden states: "`large"' state "`small"' states in sparse and structural features. Computational load or processing time of the mentioned AIM and proposed MDIS with different modes can be seen in the figure \ref{fig:time}.

\begin{figure}[!htbp]
\centering
\includegraphics[width=\columnwidth]{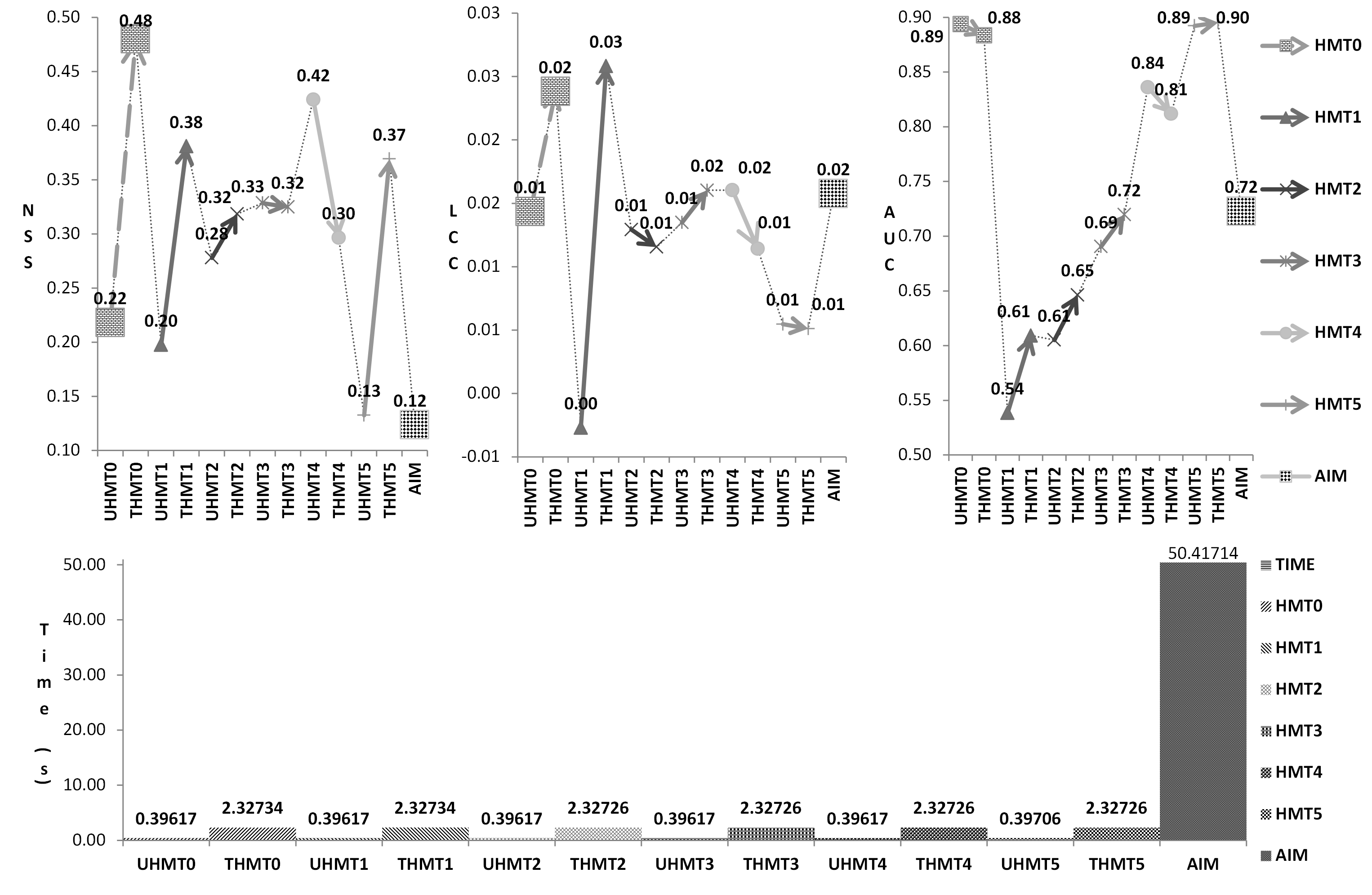}%
\caption{Summary of all MDIS against AIM}%
\label{fig:summary}%
\end{figure}

Though HMT-MDIS significantly reduces computational load for computation of information-based saliency, more verifications are necessary for their performances in terms of accuracy. We begin with evaluating two modes UHMT and THMT separately against AIM in terms of three numerical tools LCC, NSS, AUC together, Figures \ref{fig:uhmt}, \ref{fig:thmt}. Then all modes of HMTs are summarized in three plots ( the top row of Figure \ref{fig:summary} ) corresponding to three evaluating schemes NSS, LCC, AUC from left to right. Especially in the figure \ref{fig:summary}, simulation modes of the same scale level are placed next to each other for example UHMT0 is next to THMT0, UHMT1 is next to THMT1 and etc. It is intentionally arranged in that way to compare performances of different simulation modes in the same scale level. Noteworthy that, in both tables \ref{tab:uhmt} and \ref{tab:thmt} for each row is identified \textbf{maximum} values and \texttt{minimum} values by using corresponding text styles. Identification of extreme values only involves derivatives (UHMT and THMT) of MDIS modes except referenced method AIM. In the figures \ref{fig:uhmt}, \ref{fig:thmt}, extreme values are also specially marked. For example, maximum values have big solid markers while big but empty markers represent minimum points. Especially, AIM have big markers with distinguishing big cross-board texture while integrated saliency modes (U)THMT0 have small cross-board textures. These special markers help to highlight interesting facts in comparison between simulation modes of MDIS or MDIS and AIM. The same marking policy is applied for data representation in the figure \ref{fig:summary}. Meanwhile, each line in this figure has an arrow head for showing a trending direction (increasing/decreasing) when simulation mode is changed from UHMT to THMT for each scale level.

Firstly, the MDIS approach with universal parameters for each level of hidden Markov tree is analyzed in terms of accurate performance since it requires very little effort in saliency computation. Obviously it is raising a question about their accuracy in classification between center and surround class as well as generated saliency maps. According to simulation data in the table \ref{tab:uhmt} with highlighted extremum, UHMT performs pretty well against AIM in all three measurements LCC, NSS and AUC. For example, MDIS with UHMT4 mode ( 4x4 square blocks ) surpasses AIM in all measurements. It confirms validity and efficiency of our proposed methods in the information-based saliency map research field. When performances of different UHMT-MDIS modes are considered, UHMT4 with 4x4 squares  have the most consistent evaluation among all dyadic scales with maximum LCC and NSS and the second best AUC value. However, UHMT0-MDIS, integration of saliency values across scales, does not have better performance than other UHMTs except for AUC level. It shows inconsistent side of deploying HMT with predefined universal parameters while no traning effort is done for adapting the tree into image statistical mutliscale structures 

Secondly, training stage is included in the simulation of MDIS with THMT mode (Trained Hidden Markov Tree). With additional adaptivity, THMT might improves the saliency evaluation and produce more consistent results than UHMT might. This subjection is solidified by simulation data in the table \ref{tab:thmt} and they are also plotted in the figure \ref{fig:thmt}. As observed in the table, all \textbf{maximum} values locate at the THMT0 column, THMT0-MDIS over-performs AIM in all evaluating schemes. Again, the rationale of MDIS is confirmed and partly proved. Furthermore, effectiveness of traning stages are clearly shown when comparing THMT0 against UHMT0. Though AUC of THMT0 is smaller than that of UHMT0, THMT0 evaluation is better than their counterparts in both NSS and LCC scheme. This confirms usefulness of trained Hidden Markov Tree models for each sample image. In addition, the figure \ref{fig:thmt} shows supremacy of THMT0 mode, the across-scale integration mode of MDIS over other singular saliency maps at different dyadic scales in any measurement. Noteworthy, that LCC of THMT0 mode is a little bit smaller LCC of THMT1 mode; however, this small difference can be safely ignored. Comparison of UHMT-MDIS and THMT-MDIS mode-by-mode or a-dyadic-scale by a-dyadic-square between data in Table \ref{fig:uhmt} and Table \ref{fig:thmt} are shown in the figure \ref{fig:summary}. According to the figure, there are slight improvement of THMT1,THMT2 over UHMT1, UHMT2, equivalence of THMT3, UHMT3, and a reverse trend that UHMT4, UHMT5 are comparable or slightly better than THMT4, THMT5. It seems that training processes are more important when big classification windows are used. Meanwhile universal approaches of HMT work pretty well if window sizes get smaller. Two possible reasons for this observation are statistical natures of dyadic squares and characteristics of trained processes. A bigger square has richer joint-distribution of features; therefore, UHMT with fixed parameters can not marginally model that distribution well. However, parameters of THMT can be learn from analyzing images; then, significant improvement is achieved. While smaller sub-squares are less statistically distinguishing, they are successfully modeled by universal parameters of HMT. Then training processes might become redundant since UHMT would perform as well as THMT would do.

\subsection{Qualitative Evaluation}

In this section, saliency maps are analyzed qualitatively or visually. From this analysis, we want to identify (i) on which image contexts UHMT-MDIS, THMT-MDIS work well, (ii) how scale parameters affect formation of saliency maps, and (iii) how MDIS in general is compared with AIM.

\begin{figure}[!htbp]%
\subfigure{\includegraphics[width=0.49\columnwidth]{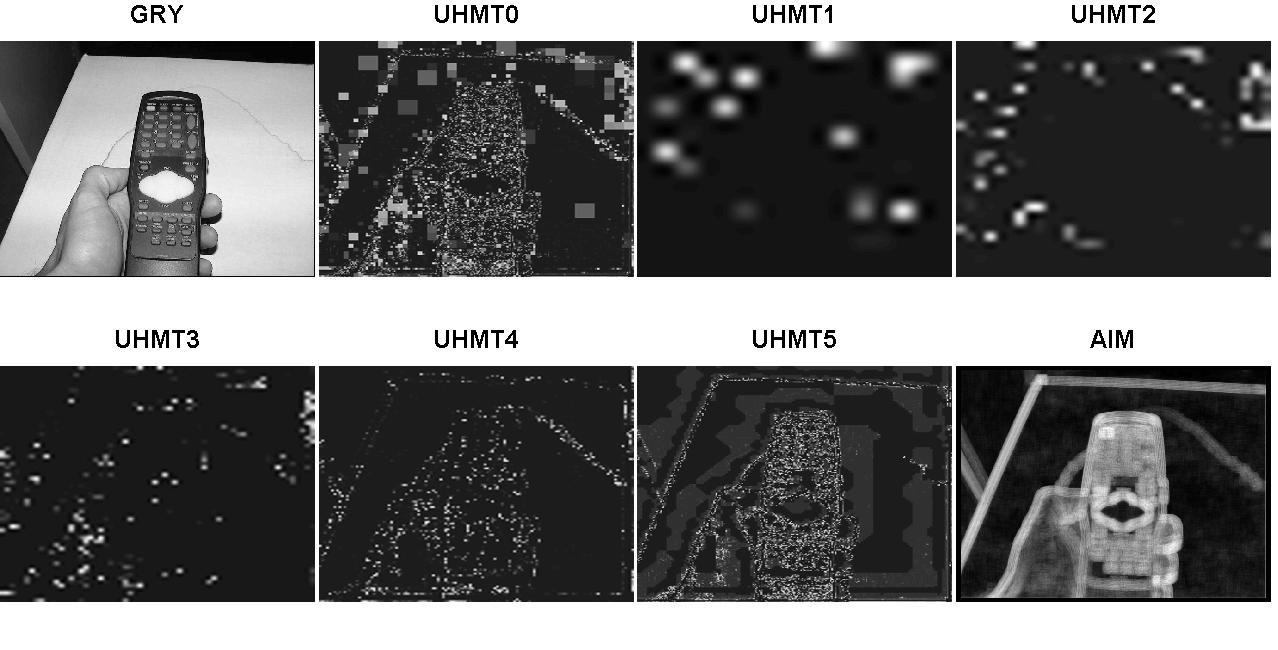}\label{fig:uhmt_19}} %
\subfigure{\includegraphics[width=0.49\columnwidth]{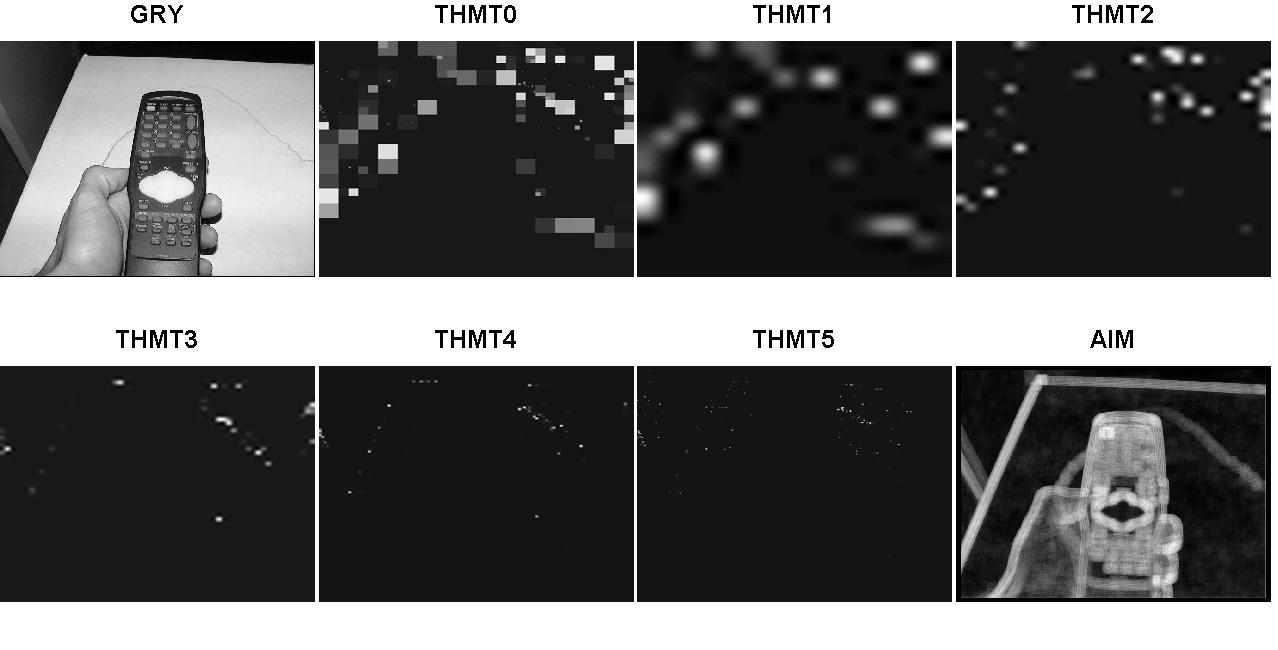}\label{fig:thmt_19}} 
\caption{Saliency Maps 1}%
\label{fig:expdis1}%
\end{figure}

In figure \ref{fig:expdis1}, the first example with central objects shows an example of good UHMT performance but bad THMT performance. All scale levels of THMT suppress features of the most obvious objects in the image center. Meanwhile, UHMT4 and UHMT5 capture significant features points of that objects; therefore, UHMT0 has much better saliency map thant THMT0. In this case, the best saliency map of MDIS approach, UHMT0, is reasonably competitive against AIM saliency map.

\begin{figure}[!htbp]%
\subfigure{\includegraphics[width=0.49\columnwidth]{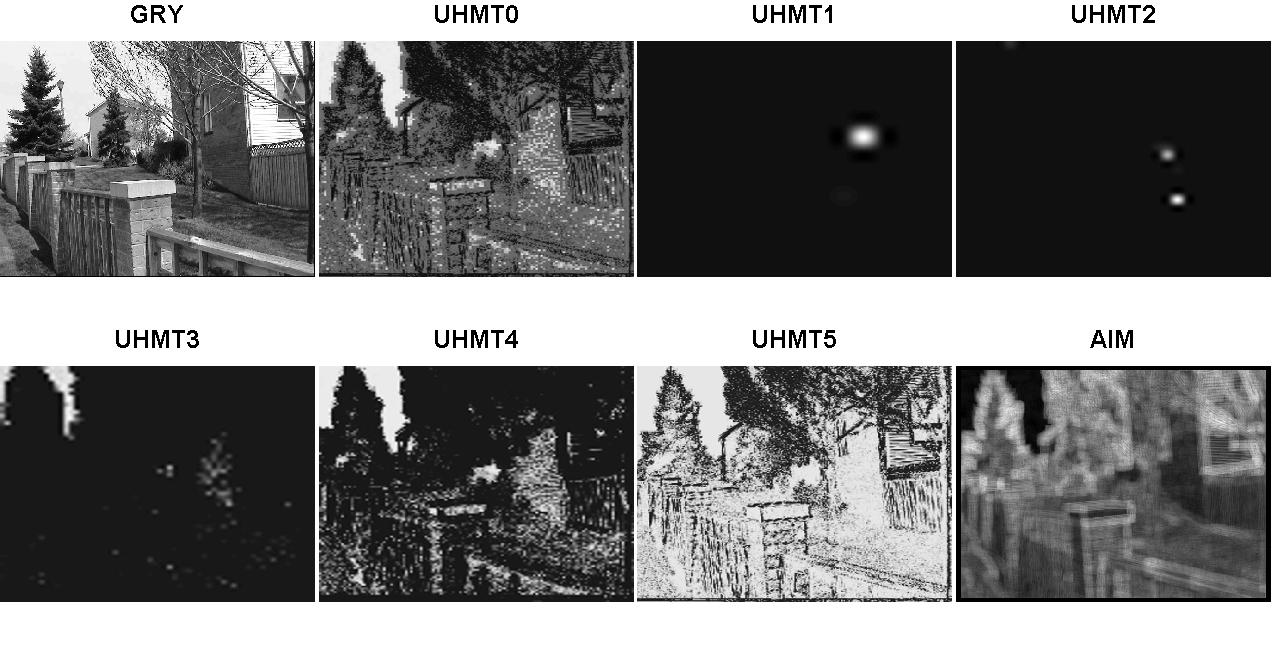}\label{fig:uhmt_102}} %
\subfigure{\includegraphics[width=0.49\columnwidth]{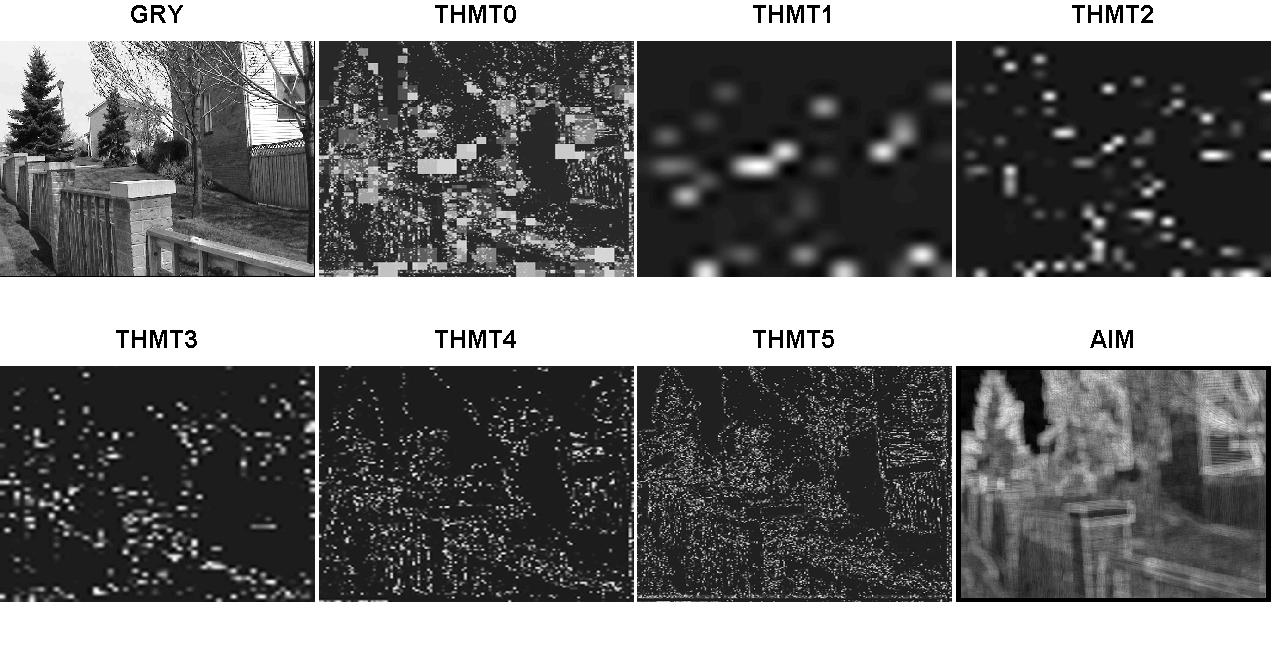}\label{fig:thmt_102}} 
\caption{Saliency Maps 2}%
\label{fig:expdis2}%
\end{figure}

In contrast with the first example, the figure \ref{fig:expdis2}, a general outdoor scene, shows an opposite situation where THMT generates more reasonable saliency maps. While UHMT0 map covers whole region of sky despite no interesting features, THMT0 correctly select available objects on the scene. Similarly, THMT does extract more meaningful features than UHMT does ( see UHMT1-5 and THMT1-5 in the figure \ref{fig:expdis2}. In addition, the best saliency map THMT0 or THMT5 highlights more discriminant features than AIM saliency map.


\begin{figure}[!htbp]%
\subfigure{\includegraphics[width=0.49\columnwidth]{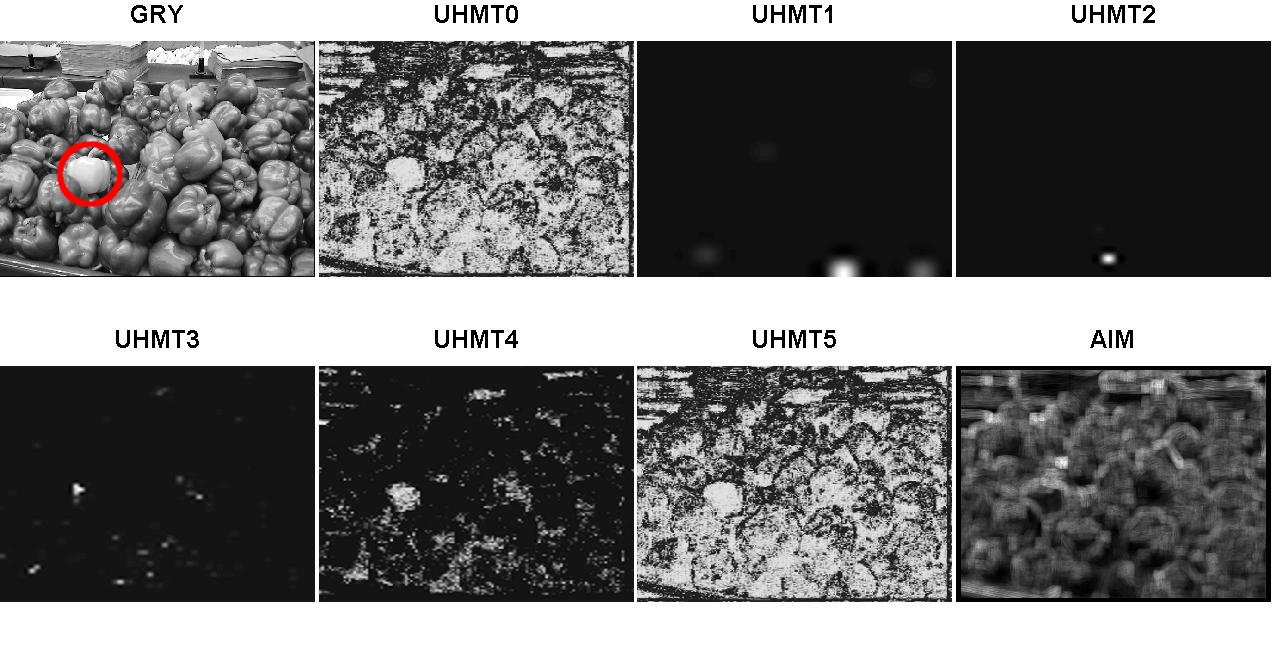}\label{fig:uhmt_111}} %
\subfigure{\includegraphics[width=0.49\columnwidth]{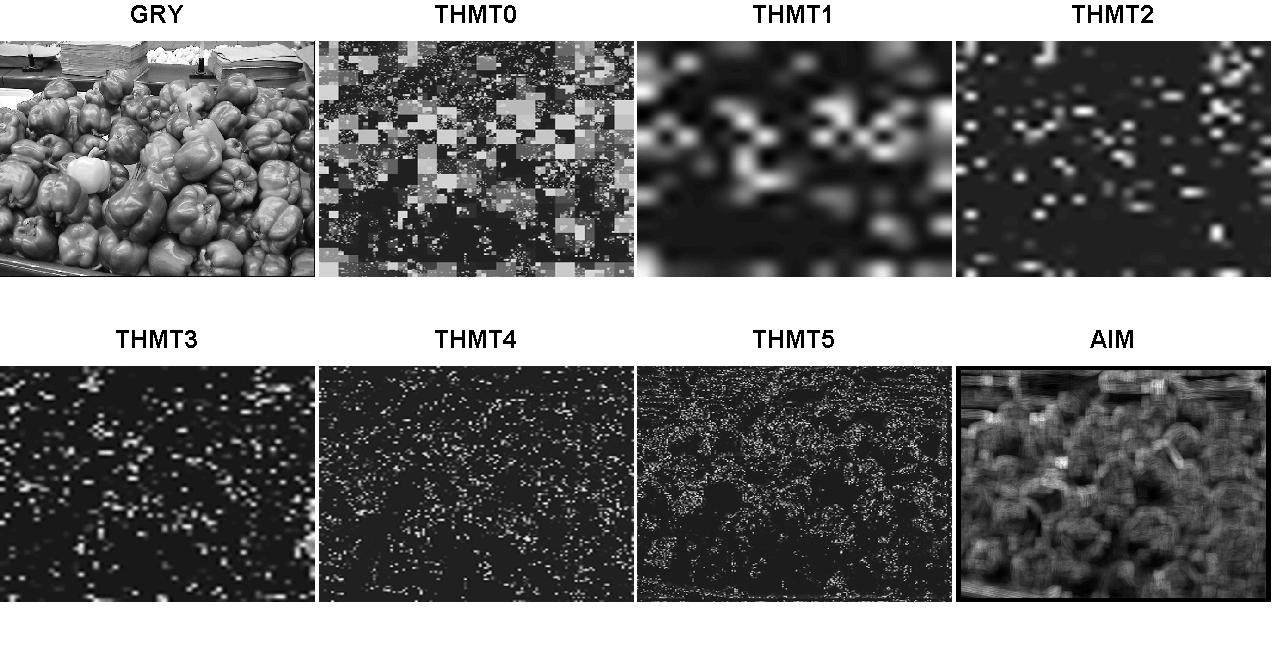}\label{fig:thmt_111}} 
\caption{Saliency Maps 3}%
\label{fig:expdis3}%
\end{figure}

The third example is chosen such that complex scenes are presented to the saliency methods. In the  figure \ref{fig:expdis3}, there are several fruits on the shelf; it is considerably complicated due to richness of edges, textures, as well as color. In general, both MDIS and AIM only partially succeed in detecting saliency regions in this image since none of them successfully high-light the fruit with different color on the shelf ( the fruit inside a red circle, Figure \ref{fig:expdis3} ). Though most MDIS with different modes and scale levels do not explicitly detect that fruit, UHMT3 and UHMT4 salency maps are able to highlight the the location of that fruit ( see UHMT3 an UHMT4 saliency maps, Figure \ref{fig:expdis3} ). The sample matches with the fact that UHMT4 data in the table \ref{tab:uhmt} has extremely good performance in all evaluation schemes. Surprisingly, there are some cases, like the figure \ref{fig:expdis3} when appropriate choices of scales and parameters of predefined HMT models can over-perform all HMT models. The interesting example of the figure \ref{fig:expdis3} opens another research direction about how HMT model can be learned and optimized; however, it is the question for another research paper.


\section{Conclusion}
\label{sec:cls}
In conclusion, multiple discriminant saliency (MDIS), an extension of DIS \cite{gao2007} under dyadic scale framework, has strong theoretical foundation as it is quantified by information-theory 
and adapted to multiple dyadic-scale structures. The performance of MDIS against AIM is simulated on the standard database with well-established numerical tools; furthermore, simulation data proves competitiveness of MDIS over AIM in both accuracy and speed. However, MDIS fails to capture salient regions in some complex scenes; therefore, the next research step is improving MDIS accuracy in such cases. In addition, implementation of MDIS algorithm in embedded systems are also considered as a possible extension.

\bibliographystyle{splncs_srt}
\bibliography{Conf2_2013}

\end{document}